# Formalization of semantic network of image constructions in electronic content

Oleg Bisikalo, Irina Kravchuk

*Abstract* — A formal theory based on a binary operator of directional associative relation is constructed in the article and an understanding of an associative normal form of image constructions is introduced. A model of a commutative semigroup, which provides a presentation of a sentence as three components of an interrogative linguistic image construction, is considered.

## I. Introduction

RECENTLY problems of computational linguistics have become especially important because of a growing demand for a natural language interface in the information technologies available to Internet users. The research is carried out in the development of an approach to modeling creative human thinking [1] and is directed towards solving the problem of increasing the level of recognition and the understanding of natural language constructions. The problems which are covered by this research are associated with support for human-computer dialogue, relevant search of information, e-learning tasks and a wide range of other problems in the realm of artificial intelligence. The purpose of this work lies in the construction of formal aids for representing an image construction as a natural language syntagma in the form of components of interrogative linguistic image constructions. We mean a syntagma as a sentence, in which only meaningful words, that conform to linguistic images, are retained, and prepositions and syncategorematic words are absent.

Notice that a different meaning of a traditional operator sign $\setminus$ (subtraction of sets) and $\oplus$ (modulo 2 addition) is given in formula expressions. In accordance with the purpose of the research they are used to signify operations of a directional relation between two images and to unite image constructions, respectively.

## II. A formal theory of a commutative semigroup of image constructions

A formal theory $Th$ is constructed as an applied theory of the first degree based on known provisions of the formal systems theory set forth in [2-4], taking into account the requirements of a concept of understanding the meaning of image constructions (IC) proposed in [5].

1. We introduce a finite alphabet consisting of symbols to be used as:
- $Al = \{A, B, ..., Z, x_1, x_2, ..., x_n, t_1, t_2, t_3\}$ – variables;
- $Con = \{\varnothing, 1, ..., n\}$ – constants;
- $\{\setminus, \oplus\}$ – symbols of binary operations defined below;
- $\{=\}$ – a binary predicate symbol "equality sign" in the sense of the set theory;
- $\{\neg, \rightarrow, \forall\}$ – logical copulas and quantifiers, where $\neg$ – negation, $\rightarrow$ – inference (if …, then …), $\forall$ – a universal quantifier;
- brackets "(", ")" and commas ",".

In accordance with a concept of understanding the meaning of image constructions [5], we consider that the symbol $\setminus$ denotes a direct relationship between two images in an associative pair $\omega \in \Omega$, whose meaning is given below, and a symbol $\oplus$ – an operation unifying image constructions «AND IC».

2. Now we will define the procedures for constructing terms (strings of characters) and formulas (acceptable expressions) of the formal theory $Th$. Terms are obtained by concatenating alphabet symbols:

$$<Term> ::= x_i j \mid x_i \in Al, j \in Con, \quad (1)$$
$$<Term> ::= <Term><Term>. \quad (2)$$

We denote terms constructed that way in the associative normal form (ANF) by the characters $t_1, t_2, t_3 \in Al$.

$$<ANF\omega> ::= x_i \setminus x_j \mid x_i, x_j \in Al; \quad (3)$$
$$<ANFterm> ::= <ANF\omega>; \quad (4)$$
$$<ANFterm> ::= <ANFterm> \oplus <ANFterm>, \quad (5)$$

where $<ANF\omega>$ is called an elementary therm in ANF.

To simplify understanding we denote individual formulas constructed in that way by characters $A, B, ..., Z \in Al$:

$$<Formula> ::= <ANFterm>, \quad (6)$$
$$<Formula> ::= (<Formula>), \quad (7)$$
$$<Formula> ::= \neg <Formula>, \quad (8)$$
$$<Formula> ::= <Formula> \rightarrow <Formula>, \quad (9)$$
$$<Formula> ::= (\forall x)<Formula> \quad (10).$$

For convenient use we add to the theory $Th$ alphabet 3 more logical connections, a quantifier $\exists$ and a functional symbol $\times$ to be used as:

$$A \,\&\, B := \neg(A \rightarrow \neg B), \quad (11)$$
$$A \vee B := \neg A \rightarrow B, \quad (12)$$
$$A \Leftrightarrow B := (A \rightarrow B) \,\&\, (B \rightarrow A), \quad (13)$$
$$(\exists x)(A) := \neg(\forall x)(\neg A), \quad (14)$$

Oleg Bisikalo – is with Vinnitsa National Technical University, Khmelnytske shose, 95, Vinnitsa, Ukraine (e-mail: obisikalo@gmail.com).
Irina Kravchuk – is with Vinnitsa National Technical University, Khmelnytske shose, 95, Vinnitsa, Ukraine (e-mail: irina.kravchuk.2010@gmail.com).

$$x_i \times x_j := (x_i \setminus x_j) \oplus (x_j \setminus x_i), \quad (15)$$

where & – a logical «AND», ∨ – a logical «OR», ⇔ – if and only if, ∃ – an existential quantifier, × – an applied functional symbol which is defined below by the symbol \. Hereafter, a formula $A$, in which a variable $x_i \in Al$ or a term $t_1$ is connected with one of the quantifiers, the formula is denoted by $A(x_i)$ or $A(t_1)$.

3. We select a set of formulas that are considered axiom schemes.

Logical axioms (3.1÷3.3 – expressions calculus, 3.4÷3.5 – first-order predicate calculus [3]):

3.1. $A \to (B \to A)$.

3.2. $(A \to (B \to C)) \to ((A \to B) \to (A \to C))$.

3.3. $(\neg B \to \neg A) \to ((\neg B \to A) \to B)$.

3.4. $\forall x_i A(x_i) \to A(t_1)$ [where $A(x_i)$ is a formula from $Th$ and $t_1$ is a term from $Th$, free for $x_i$ in $A(x_i)$].

3.5. $\forall x_i (A \to B) \to (A \to \forall x_i B)$ [if a formula A does not include free occurrences $x_i$].

Proper axioms (3.6÷3.11 – axioms of a commutative semigroup [4], 3.12÷3.14 – applied axioms (products) of the theory):

3.6. $\forall t_1 \forall t_2 \forall t_3 (t_1 \oplus (t_2 \oplus t_3) = (t_1 \oplus t_2) \oplus t_3)$ (associativity).

3.7. $\forall t_1 (t_1 = t_1)$ (reflectiveness).

3.8. $\forall t_1 \forall t_2 (t_1 = t_2 \to t_2 = t_1)$ (symmetry).

3.9. $\forall t_1 \forall t_2 \forall t_3 (t_1 = t_2 \to (t_2 = t_3 \to t_1 = t_3))$ (transitivity).

3.10. $\forall t_1 \forall t_2 \forall t_3 (t_2 = t_3 \to (t_1 \oplus t_2 = t_1 \oplus t_3) \& (t_2 \oplus t_1 = t_3 \oplus t_1))$ (substitution).

3.11. $\forall t_1 \forall t_2 (t_1 \oplus t_2 = t_2 \oplus t_1)$ (commutativity).

3.12. $\forall x_i, x_j, x_k (x_i j x_k \to x_j \setminus x_i \oplus x_k)$ (transformation of a string to terms in ANF).

3.13. $\forall x_i, x_j (x_i j \to x_j \setminus x_i)$ (finite transformation of a string to a term in ANF).

3.14. $\forall x_i, x_j (x_i \setminus x_j \oplus x_i \setminus x_j \to x_i \setminus x_j)$ (reduction of a term in ANF)

4. We define a finite set of rules of inference, which provide a different set of formulas from some finite set of formulas.

$A, A \to B \mapsto B$ «Modus ponens»,

$A \mapsto (\forall t) A$ «a generalization rule»,

where the notation $\tilde{A} \mapsto \grave{A}$ means that A is a result of the formulas set Γ.

Besides theorems of the formal theory of the first-order predicates, in the theory $Th$ such proper theorems are true.

Theorem 1. $<Term> \to <ANFterm>$.

Proving by induction on a length of derivation $B_1, B_2, ..., B_k = B$:

a) $<Term>$ – a hypothesis;

b) $x_1 j$ – an induction base: according to the 1st definition of a term (2a);

c) $x_j \setminus x_1$ – 3.13 before b);

d) $<ANFterm>$ – according to the 1st definition of a term in ANF;

e) $x_1 j x_2 i$ – or according to the 2nd definition of a term;

f) $x_j \setminus x_1 \oplus x_2 i$ – 3.12 before e);

g) $x_j \setminus x_1 \oplus x_i \setminus x_2$ – 3.13 before f);

h) $<ANFterm>$ – according to the 2nd definition of a term in ANF;

i) $\underbrace{x_1 j x_2 i ... x_k}_{k-1} l$ – induction transfer: according to the 2nd definition of a term;

j) $<ANFterm> \oplus x_k l$ – 3.12 before i) k-1 times;

k) $<ANFterm> \oplus x_l \setminus x_k$ k) – 3.13 before j);

l) $<ANFterm>$ – according to the 2nd definition of a term in ANF.

Theorem 2. A similar proof of such a theorem:
$<ANFterm> \to <ANFq> \oplus <ANF?> \oplus <ANFa>$,

where $<ANF\omega> = x_i \setminus x_j \mid x_i, x_j \in Al$ for convenience is denoted by $<ANF?>$;

$<ANFa>$ – all elementary terms of $<ANFterm>$, where a symbol $x_j$ is the first, then the next symbol is substituted recursively based on the principle of depth-first search, but if $<ANF?> = x_j \setminus x_i$ is found, then a symbol $x_i$ and all symbols following it are not taken into account;

$<ANFq>$ – all other elementary terms that compose $<ANFterm>$.

III. A MODEL OF A COMMUTATIVE SEMIGROUP OF IMAGE CONSTRUCTIONS AND EXAMPLES

We will now consider a model of the formal theory $Th$ as a commutative semigroup of image constructions. Within the model we consider that function symbols denote the following relations between two linguistic image [5]: \ – «principal-subordinate» relation, × – «subject-predicate» relation. Under the term we understand the image construction of a simple sentence (syntagma), and under the formula of the theory – an image analog of a logical natural language expression. We denote individual images from the set $I = \{x_1, x_2, ..., x_n\}$ by the characters $x_1, x_2, ..., x_n$, terms in ANF – by characters $t_1, t_2, t_3$, formulas – $A, B, ..., X$, an unknown subject – $Y$, an unknown predicate – $Z$. The elementary term in ANF $<ANF\omega> \mid <ANF?>$ is called an associative pair of images, where │ – a denotation of the OR operator in Backus-Naur Form. Terms or image constructions are constructed from natural language sentences based on this rule 1: a sentence of $k$ words is written as a string of $2 \cdot k$ characters, where each $i$-th word in a sentence is put in correspondence to a linguistic image $x_i \in Al$, and after it $j \in Con$ is recorded as a indicator of another image $x_j$ of this sentence that is principal to a subordinate image $x_i$. If homogeneous parts are found in a sentence, then the possible cases are

$$(x_1 \& x_2)j \to x_j \setminus x_1 \oplus x_j \setminus x_2 \quad (16)$$

or

$$(x_1 \& x_2)j \oplus <АНФтерм> \oplus x_1 \setminus x_j \to x_j \setminus x_1$$
$$\oplus x_j \setminus x_2 \oplus <АНФтерм> \oplus x_1 \setminus x_j \oplus x_2 \setminus x_j \quad (17)$$

Limitations of the considered model:

• natural language sentences must have both subject and predicate, otherwise they are included artificially using $Y$ and/or $Z$ symbols;

• rule 1 applies only to meaningful words in a sentence that correspond to image constructions, and punctuation marks, prepositions and syncategorematic words in sentences are not accounted for.

Within the model, theorems of the formal theory $Th$ receive this interpretation.

Theorem 1. Any term that corresponds to a natural language sentence (syntagma) and is based on rule 1, can be represented as a term in ANF:
$$<Term> \to <ANFterm>. \quad (18)$$

Theorem 2. If from a sentence represented in the form of a term in ANF $<ANFterm>$ one selects one associative pair as an interrogative pronoun, then all elementary terms that directly dependent on this pair in ANF will make an answer, and all other elementary terms from $<ANFterm>$ – an interrogative sentence:
$$<ANFterm> \to <ANFq> \oplus <ANF?> \oplus <ANFa>. \quad (19)$$

For convenient use of the model of the formal theory $Th$ in content elements we introduce rule 2:

$<ANFterm> \to <\underline{ANF?}><tQ>?<tA>$,

where

$<tQ> := (x_i \mid <ANFq> = \varnothing) \mid$
$(x_i x_l ... x_m x_k \mid <ANDFq> = x_i \setminus x_l \oplus ... \oplus x_m \setminus x_k)$;

$<tA> := (x_j \mid <ANFa> = \varnothing) \mid$
$(x_j x_l ... x_m x_k \mid <ANFa> = x_j \setminus x_l \oplus ... \oplus x_m \setminus x_k)$;

? – an additional sign that denotes the end of a interrogative part $<ANFterm>$.

Strings of characters $x_i x_l ... x_m x_k$ received for $<tQ>$ and $<tA>$ are rewritten by removing those characters from left to right, which recur. Formally, for 2ns symbol $x_1 x_2 \to ([x_2 = x_1] x_1, x_1 x_2)$, for $k$-th symbol $x_1 x_2 ... x_k \to ([x_k = x_1 \mid x_k = x_2 \mid ... \mid x_k = x_{k-1}] x_1 x_2 ... x_{k-1}, x_1 x_2 ... x_k)$.

To demonstrate the capabilities of the model of IC commutative semigroup of the formal theory $Th$ we consider examples of sentences in English and Russian.

Example 1. Once I saw (a) little bird ($x_1 x_2 x_3 x_4 x_5$).

According to rule 1, we construct a term
$x_1 3 x_2 3 x_3 2 x_4 5 x_5 3$;

– a product 3.12 to substring $x_1 3 x_2$ leads to
$x_3 \setminus x_1 \oplus x_2 3 x_3 2 x_4 5 x_5 3$;

– a product 3.12 to substring $x_2 3 x_3$ leads to
$x_3 \setminus x_1 \oplus x_3 \setminus x_2 \oplus x_3 2 x_4 5 x_5 3$;

– a product 3.12 to substring $x_3 2 x_4$ leads to
$x_3 \setminus x_1 \oplus x_3 \setminus x_2 \oplus x_2 \setminus x_3 \oplus x_4 5 x_5 3$;

– a product 3.12 to substring $x_4 5 x_5$ leads to

$x_3 \setminus x_1 \oplus x_3 \setminus x_2 \oplus x_2 \setminus x_3 \oplus x_5 \setminus x_4 \oplus x_5 3$;

– a product 3.13 to substring $x_5 3$ leads to
$x_3 \setminus x_1 \oplus x_3 \setminus x_2 \oplus x_2 \setminus x_3 \oplus x_5 \setminus x_4 \oplus x_3 \setminus x_5$ – we have a term in ANF.

Thus, an initial natural language construction in ANF is as follows:

saw \ once $\oplus$
saw \ I $\oplus$
I \ saw $\oplus$
bird \ little $\oplus$
saw \ bird.

We denote $<ANF?> := x_3 \setminus x_1$ by a word <when?>. According to theorem 2, $<ANFa> \to \varnothing$, $<ANFq> \to x_3 \setminus x_2 \oplus x_2 \setminus x_3 \oplus x_3 \setminus x_5 \oplus x_5 \setminus x_4$.

Then, according to rule 2, $<tA> \to x_1$, and $<tQ> \to x_3 x_2 x_5 x_4$. Thus, we have the following result:

when? saw I bird little ? once.

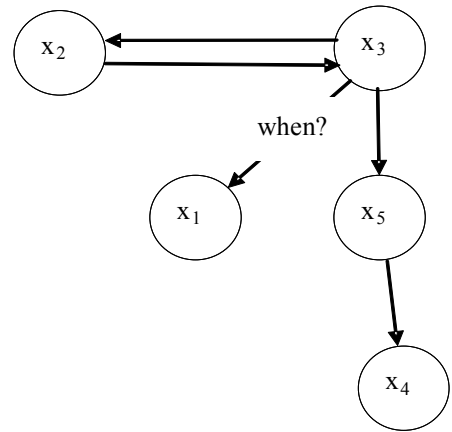

Fig. 1. Graph of the therm (sentence) $x_1 3 x_2 3 x_3 2 x_4 5 x_5 3$ with selection of the associative pair $x_3 \setminus x_1$

It is easy to prove an equivalence of a graph model [6] and presented in this paper theory $Th$ that is a subject of the further research from an applied point of view. This fact allows us to use known search algorithms on graphs for solving applied problems of finding the optimal path, the traversal of the graph and search during processing natural language constructions. Presented in Fig. 1 is the graph of a sentence illustrating an example of using of the theory $Th$ model. The following notation is used:

O – a linguistic image – a part of a sentence;

$\to$ – relation between principal and subordinate members of a sentence;

$\xrightarrow{word?}$ – an interrogative pronoun of an associative pair that is used to form an interrogative sentence.

Example 2 (Russian). Забытую песню несет ветерок (в) задумчивых травах звеня ($x_1 x_2 x_3 x_4 x_5 x_6 x_7$).

According to rule 1, we construct a term
$x_1 2 x_2 3 x_3 4 x_4 3 x_5 6 x_6 7 x_7 3$;

– a product 3.12 to substring $x_1 2 x_2$ leads to

$x_2 \backslash x_1 \oplus x_2 3 x_3 4 x_4 3 x_5 6 x_6 7 x_7 3$ ;

– a product 3.12 to substring $x_2 3 x_3$ leads to
$x_2 \backslash x_1 \oplus x_3 \backslash x_2 \oplus x_3 4 x_4 3 x_5 6 x_6 7 x_7 3$ ;

– a product 3.12 to substring $x_3 4 x_4$ leads to
$x_2 \backslash x_1 \oplus x_3 \backslash x_2 \oplus x_4 \backslash x_3 \oplus x_4 3 x_5 6 x_6 7 x_7 3$ ;

– a product 3.12 to substring $x_4 3 x_5$ leads to
$x_2 \backslash x_1 \oplus x_3 \backslash x_2 \oplus x_4 \backslash x_3 \oplus x_3 \backslash x_4 \oplus x_5 6 x_6 7 x_7 3$ ;

– a product 3.12 to substring $x_5 6 x_6$ leads to
$x_2 \backslash x_1 \oplus x_3 \backslash x_2 \oplus x_4 \backslash x_3 \oplus x_3 \backslash x_4 \oplus x_6 \backslash x_5 \oplus x_6 7 x_7 3$ ;

– a product 3.12 to substring $x_6 7 x_7$ leads to
$x_2 \backslash x_1 \oplus x_3 \backslash x_2 \oplus x_4 \backslash x_3 \oplus x_3 \backslash x_4 \oplus x_6 \backslash x_5 \oplus x_7 \backslash x_6 \oplus x_7 3$ ;

– a product 3.12 to substring $x_7 3$ leads to
$x_2 \backslash x_1 \oplus x_3 \backslash x_2 \oplus x_4 \backslash x_3 \oplus x_3 \backslash x_4 \oplus x_6 \backslash x_5 \oplus x_7 \backslash x_6 \oplus x_3 \backslash x_7$

– we have a term in ANF.

Thus, an initial natural language construction in ANF is as follows:

песню \ забытую $\oplus$

несет \ песню $\oplus$

ветерок \ несет $\oplus$

несет \ ветерок $\oplus$

травах \ задумчивых $\oplus$

звеня \ травах $\oplus$

несет \ звеня .

We denote $<ANF?>:= x_3 \backslash x_2$ by a word <what?>. According to theorem 2, $<ANFa> \to x_2 \backslash x_1$, $<ANFq> \to x_4 \backslash x_3 \oplus x_3 \backslash x_4 \oplus x_6 \backslash x_5 \oplus x_7 \backslash x_6 \oplus x_3 \backslash x_7$.

Thus, according to rule 2, $<tA> \to x_2 x_1$, and $<tQ> \to x_3 x_4 x_7 x_6 x_5$. Thus, we have the following result:

what? несет ветерок звеня задумчивых травах ? песню забытую.

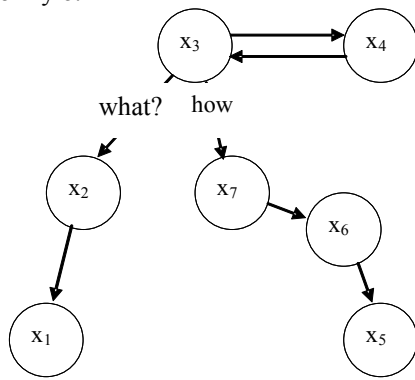

Fig. 2. Graph of the therm $x_1 2 x_2 3 x_3 4 x_4 3 x_5 6 x_6 7 x_7 3$ with selection of the associative pair $x_3 \backslash x_2$ and $x_3 \backslash x_7$

Now we denote $<ANF?>:= x_3 \backslash x_7$ by a word <how?>. According to theorem 2, $<ANFa> \to x_7 \backslash x_6 \oplus x_6 \backslash x_5$, $<ANFq> \to x_4 \backslash x_3 \oplus x_3 \backslash x_4 \oplus x_3 \backslash x_2 \oplus x_2 \backslash x_1$.

Thus, according to rule 2, $<tA> \to x_7 x_6 x_5$, and $<tQ> \to x_3 x_4 x_2 x_1$. Thus, we have the following result:

how? несет ветерок песню забытую ? звеня задумчивых травах.

Fig. 2 shows the graph of a sentence with selection of two associative pairs.

## IV. Conclusion

Thus, the given examples demonstrate the intuitive intelligibility of the results of applying the model of IC commutative semigroups of the formal theory $Th$ to natural language structures in the form of sentences in English and Russian. Unlike existing formal theories, a binary operator of directional associative relation and the concept of ANF, according to the concept of understanding the sense of an electronic text content, are applied in the formal theory $Th$. A model of image constructions commutative semigroup that, based on the theory $Th$, provides a representation of IC of a natural language syntagma as 3 components of an interrogative construction of linguistic images.